# When Symptoms Are Not Enough: Evidence-Weighting Patterns in Large Language Model Psychiatric Screening

Jianfeng Zhu[1], Megan Korhummel[2] , Ruoming Jin[1] & Karin G. Coifman[2]
Departments of Computer Science[1] and Psychological Science[2]
Kent State University
Kent, OH USA

**Abstract**
As demand for mental health care outpaces clinician-delivered assessment, scalable screening tools are increasingly needed. Large language models (LLMs) may identify psychiatric risk from patient narratives, yet their reliability across diagnoses, demographic subgroups, and clinically relevant evidence-use patterns remains uncertain. Here, we introduce a SCID-anchored benchmark of 555 semi-structured experiential interviews paired with SCID-derived diagnostic reference labels for anxiety disorder, major depressive disorder, posttraumatic stress disorder, and any current mental health disorder. Using zero-shot task-specific prompting, we evaluated multiple state-of-the-art LLMs and analyzed whether false-negative errors reflected missed psychiatric evidence or differential weighting of symptom, impairment, and protective-context cues. Model performance varied substantially across tasks and model families, with accuracy ranging from 0.49 to 0.86 and Matthews correlation coefficient values ranging from 0.16 to 0.38. GPT-4.1 Mini and GPT-5 Mini showed the most consistent disorder-specific accuracy. Subgroup analyses identified higher depression-classification accuracy among male than female participants, no consistent age-related pattern, and modest, non-uniform variation across race strata. Output-rationale pattern analyses showed that false-negative anxiety and PTSD classifications often contained explicit symptom evidence in the model-generated rationale text, but were also accompanied by protective-context language such as preserved functioning, coping ability, and social support. These are output-level patterns in model-generated rationales and should not be interpreted as direct evidence of internal model reasoning processes. Functional-impairment language shifted model outputs toward positive classifications, whereas protective-context language shifted outputs away from them. These findings suggest that LLMs may support scalable psychiatric screening at a research stage, but their tendency to discount symptom evidence in the presence of protective-context language requires prospective clinical validation before deployment.

## 1 Introduction

Mental health disorders impose a substantial global burden, yet access to clinician-delivered assessment remains limited [1,2]. With only about 2% of global health budgets allocated to mental health, the majority of people living with mental disorders reside in settings with severe shortages of trained mental health professionals [3,4]. Existing screening and triage systems often depend on self-report questionnaires, structured interviews, or clinician-intensive evaluations, which are difficult to scale in settings with limited clinical resources [5,6]. This gap has motivated interest in language-based approaches that can extract mental health–relevant signals from patient narratives and support earlier identification of psychiatric risk. In this context, scalable screening tools are needed not to replace clinician judgment, but to help identify individuals who may require further clinical evaluation and to reduce the risk that clinically meaningful distress is missed.

Language is central to psychiatric assessment because patients often communicate symptoms, impairment, coping, distress, and contextual stressors through narrative accounts [7,8]. Prior Natural Language Processing (NLP) studies have shown that written language, speech transcripts, online posts, and clinical narratives contain signals associated with depression, anxiety, PTSD, suicidality, and broader psychological distress [9–13]. However, much of this work relies on self-disclosed labels, online behavioral data, single-condition tasks, or symptom-specific datasets, limiting its relevance to clinically anchored psychiatric screening.

Recent advances in large language models (LLMs) have expanded the possibility of using patient narratives for mental health assessment [14]. Unlike traditional supervised models that require task-specific training data, LLMs can be prompted to interpret long-form narratives, produce clinically framed judgments, and provide natural-language rationales for their decisions [10,15–17]. Prior studies suggest that LLMs may support symptom screening, diagnostic reasoning, and clinician-facing decision support [18,19], However, work remains focused on final classification performance, single-disorder screening, or narrowly defined symptom detection, rather than evaluating how LLMs behave across multiple potentially co-occurring psychiatric conditions. Most evaluations focus on final classification performance, benchmark accuracy, or the apparent quality of model-generated explanations. Less is known about how LLMs weigh different forms of clinically relevant evidence when producing false-negative psychiatric screening decision [20]. Several gaps remain critical for clinical translation. First, few benchmarks evaluate LLM-based screening using semi-structured experiential interviews paired with SCID-derived diagnostic reference labels. Second, subgroup variation in model performance remains insufficiently characterized, particularly across sex, age, and race strata. Third, most studies emphasize final classification accuracy, leaving unclear how models use clinically relevant evidence when producing errors. In psychiatric screening, false-negative classifications may occur not because symptom evidence is absent, but because models weigh symptoms against contextual cues such as preserved functioning, coping capacity, social support, or diagnostic uncertainty.

To address these gaps, we introduce a SCID-anchored benchmark of 555 (Mage = 39.4; sd = 16.5 ; 50.1%female ; 77.8%white) semi-structured experiential interviews across a range of samples and stressful circumstances. Participants also completed structured clinical interviews for the diagnosis of DSM-IV-TR/DSM-5 disorders [21]. Interviews were conducted by MA-level clinicians who achieve high reliability at the symptom and diagnostic level [22]. We evaluate multiple state-of-the-art LLMs under zero-shot task-specific prompting and assessed diagnostic performance using accuracy, precision, recall, specificity, F1 score, and Matthews correlation coefficient. We further examined subgroup variation by sex, age, and race, and analyzed model-generated inference narratives to determine whether false-negative errors reflected missed psychiatric evidence or differential weighting of symptom, functional-impairment, and protective-context cues.

**Our research addresses three interrelated questions:**
**RQ1:** How accurately can state-of-the-art LLMs perform zero-shot mental health screening from semi-structured experiential interviews using SCID-derived diagnostic reference labels?
**RQ2:** How does screening performance vary across diagnostic categories and demographic subgroups, including sex, age, and race?
**RQ3:** Do false-negative model errors reflect failure to detect psychiatric symptom evidence, or differential weighting of symptom, functional-impairment, and protective-context cues in model-generated inference narratives?

## 2 Methods

### 2.1 Data Sources, Reference Labels, and Screening Tasks

The dataset included 555 participants who completed semi-structured interviews focused on recent stressful experiences. The sample included 278 females (50.1%) and 276 males (49.7%), and one participant with missing sex information (0.2%). The majority of participants were aged 18–44 years (n = 346, 62.3%), followed by 45–64 years (n = 167, 30.1%), and 65 years or older (n = 40, 7.2%). The interviews did not directly ask participants to report psychiatric symptoms or distress. Interview modules asked participants to describe stressful experiences, emotional responses, coping behaviors, and perceived social support; the five interview questions and example responses are provided in Appendix A, with additional dataset documentation available on OSF [23].
In parallel with the interview assessment, participants completed a structured psychiatric evaluation using the Structured Clinical Interview for DSM-IV or DSM-5 (SCID)[24,25]. SCID-derived diagnoses were used as reference labels for model benchmarking. Diagnostic variables were coded as binary indicators (0 = absent, 1 = present), with values coded as 999 treated as missing. Four screening outcomes were defined from these labels: anxiety disorder, depression, PTSD, and any current mental health disorder (MHD). Anxiety disorder was coded as positive if any of GAD, SAD, PD, or OCD was present; depression, PTSD, and MHD were defined using the MDE, PTSD, and AnyDiagnosis variables, respectively.

### 2.2 LLM Screening and Evaluation

#### *2.2.1 Models and Prompting Framework*

Five large language models were evaluated: LLaMA 3, DeepSeek, GPT-4o Mini, GPT-4.1 Mini, and GPT-5 Mini. These models were queried through API-based chat completion calls, and inference parameters followed provider defaults because they were not explicitly set in the original script. All models were evaluated using zero-shot, task-specific prompting, in which models were instructed to perform the screening task without labeled examples or additional fine-tuning [26].
Each task-specific prompt asked the model to evaluate one psychiatric condition at a time from the participant transcript. For each transcript, the model generated a brief clinical-style inference rationale summarizing symptom-related and contextual evidence, followed by a binary screening prediction (“Yes” or “No”). This prompting framework was used across all primary performance evaluations. Full task-specific prompts and example model outputs are provided in Appendix A.

#### *2.2.2 Screening Performance Evaluation*

Model screening performance was evaluated using binary classification metrics, including accuracy, precision, recall, specificity, F1 score, and Matthews correlation coefficient (MCC). Precision was defined as positive predictive value, and recall was defined as sensitivity. Metrics were computed separately for each model and screening task.

Differences in accuracy across models were evaluated using Cochran's Q tests for related proportions because all models were applied to the same participant transcripts [27]. When omnibus tests were significant, pairwise model comparisons were conducted using McNemar tests with Bonferroni correction [28]. Statistical comparisons were performed separately for each screening task.

*2.2.3 Demographic Subgroup Analysis*

To characterize demographic variability in screening performance, the same classification metrics were recomputed within sex, age, and race/ethnicity strata. Sex was categorized as female or male. Age was grouped into 18–44, 45–64, and 65+ years [29,30]. Race/ethnicity was dichotomized as White and Non-White because smaller racial/ethnic groups had limited representation. Subgroup analyses were conducted separately for each model and screening task and treated as descriptive analyses.

## 2.3 Psychiatric Evidence Integration Analysis

To examine how output-rational text patterns differed between true-positive and false-negative classifications, we conducted a post hoc evidence integration analysis of GPT-4.1 Mini inference texts. GPT-4.1 Mini was selected because it produced concise, consistently structured rationales across all four tasks and showed strong overall benchmark performance. This analysis characterizes patterns in model-generated output text and should not be interpreted as direct evidence of internal model cognition or causal access to the model decision process. Evidence features were extracted from the model-generated rationale rather than from the original participant transcript.

Three categories of evidence features were extracted from each rationale using rule-based regular-expression dictionaries  symptom evidence, functional impairment, and protective context [31–33]. Symptom evidence captured task-specific psychiatric symptom language, derived from validated screening domains, including PHQ-9 domains  [34], GAD-7 [35], and PCL-5 [36]. Functional impairment captured language indicating disruption in occupational, social, or daily functioning. Protective context included preserved functioning, active coping, and social support language. Feature counts reflected the number of matched keyword patterns within each rationale. The full keyword list is provided in Appendix C.

Binary screening predictions were categorized as true positive (TP), false positive (FP), false negative (FN), or true negative (TN) relative to SCID-derived reference labels. Among reference-positive cases, evidence feature counts were compared between TP and FN classifications using two-sided Mann–Whitney U tests [37], with rank-biserial correlation reported as the effect-size estimate [38]. Multiple comparisons across the 12 task-feature combinations (4 tasks × 3 features) were controlled using the Benjamini–Hochberg false discovery rate procedure[39].

To further characterize the linguistic basis of GPT-4.1 Mini's binary output, standardized logistic regression models were fitted separately for each screening task with the three evidence feature counts as predictors. Two outcome specifications were estimated in parallel: GPT-4.1 Mini's binary output, and the SCID-derived reference label. The SCID model served as an external grounding check to assess whether the same evidence domains were associated with reference-confirmed diagnostic status. Model-level feature contributions were estimated from the fitted logistic models to summarize how each evidence domain contributed to model-positive and SCID-positive classifications [40].

## 3 Results

Among the 555 participants, SCID-derived reference labels indicated anxiety disorder in 150 participants (27.0%), major depressive disorder in 81 (14.6%), PTSD in 125 (22.5%), and any current mental health disorder in 286 (51.7%).

### 3.1.1 Primary Screening Performance under Task-Specific Prompting

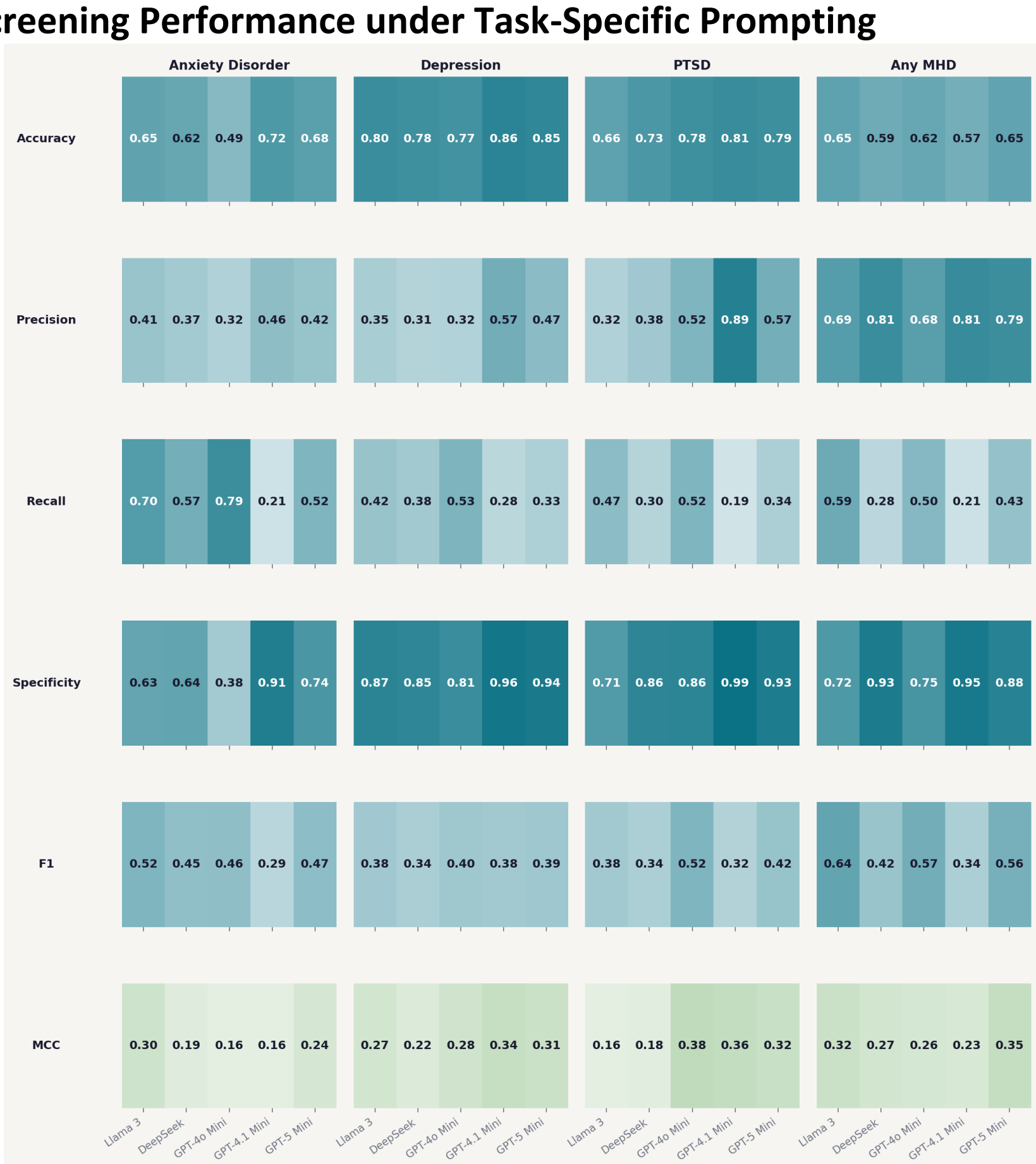

Figure 1. Model Performance Across Mental Health Disorders

Figure 1 summarizes screening performance across five LLMs and four diagnostic tasks. Accuracy varied across both models and disorders, ranging from 0.49 to 0.86 across model–task combinations. Accuracy differed significantly across models for all four screening tasks, including anxiety disorder, depression, PTSD, and any current mental health disorder (all Cochran's Q tests, $p < .001$). Variation was largest for anxiety and PTSD classification. GPT-4o Mini showed the lowest accuracy for anxiety screening (0.49), whereas GPT-4.1 Mini achieved the highest accuracy for anxiety (0.72), depression (0.86), and PTSD (0.81). For any current mental health disorder, LLaMA3 and GPT-5 Mini showed the highest accuracy (0.65). Post hoc McNemar tests with Bonferroni correction identified significant pairwise differences between models; full pairwise results are provided in Appendix B.

### 3.1.2 Demographic Variation in Screening Performance

Figure 2 summarizes model accuracy across diagnostic tasks stratified by sex. Depression classification showed the clearest sex-related pattern: accuracy was consistently higher among male participants than female participants across all five models. The largest differences were observed for GPT-4.1 Mini (male: 0.91; female: 0.82) and GPT-5 Mini (male: 0.87; female: 0.82), while smaller differences were observed for GPT-4o Mini (male: 0.79; female: 0.74), LLaMA3 (male: 0.89; female: 0.72), and DeepSeek(male: 0.85; female: 0.72). Accuracy differences by sex were smaller and less consistent for anxiety disorder, PTSD, and any current mental health disorder.

Within sex strata, Cochran's Q tests indicated significant model-level differences for most diagnostic tasks, with the exception of any current mental health disorder among male participants. Full subgroup-specific omnibus and pairwise McNemar test results are provided in Appendix B. Across both male and female participants, GPT-4.1 Mini generally showed the highest or near-highest accuracy for anxiety, depression, and PTSD, whereas LLaMA3 and GPT-5 Mini performed comparatively well for any current mental health disorder.

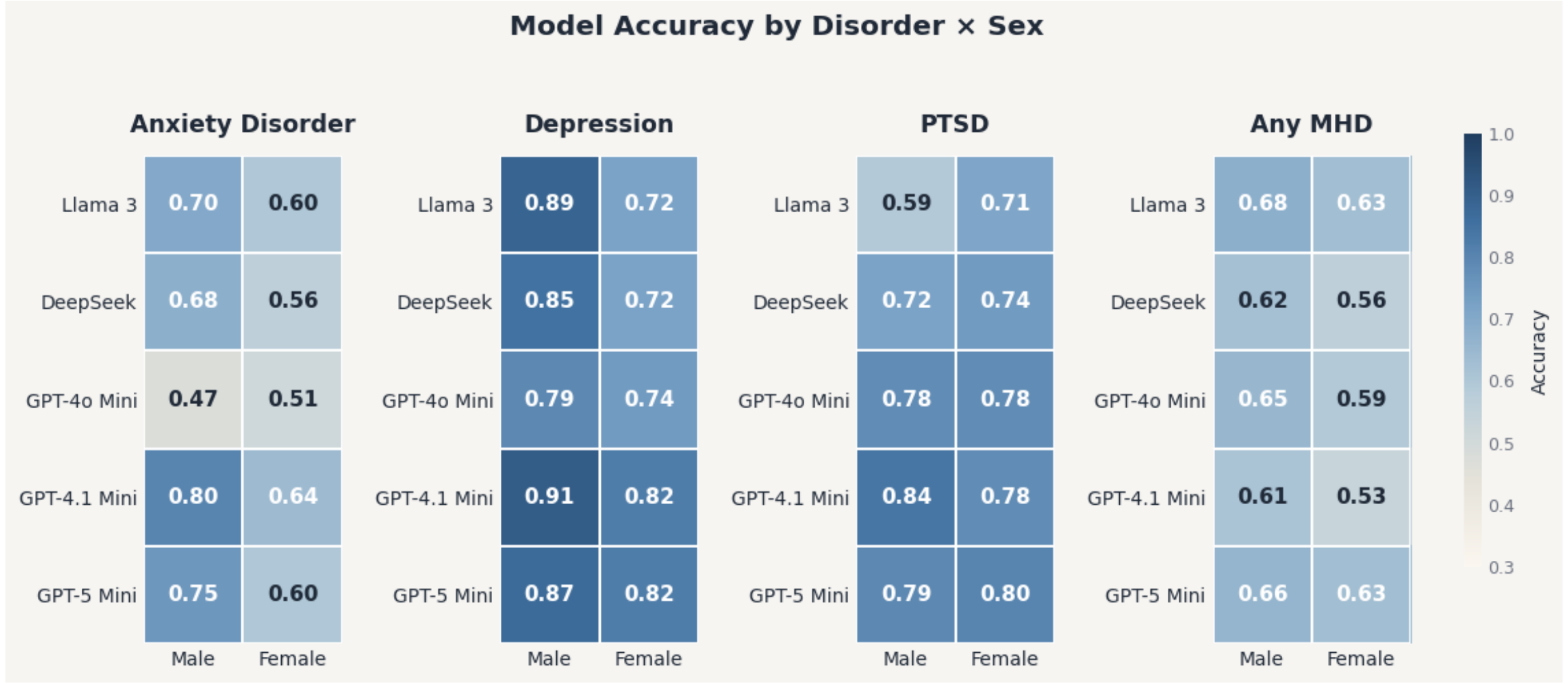

**Figure 2.** Sex-Stratified Model Accuracy Across Mental Health Screening Tasks

Figure 3 summarizes model accuracy across mental health screening tasks stratified by age group. Across model–task–age combinations, accuracy ranged from 0.45 to 0.93, indicating variability across both age strata and diagnostic tasks. Depression classification showed relatively stable accuracy across age groups, with GPT-4.1 Mini reaching 0.86, 0.87, and 0.93 in the 18–44, 45–64, and 65+ groups, respectively. PTSD performance was also highest or near-highest for GPT-4.1 Mini across age groups, whereas anxiety and any current mental health disorder showed less consistent model rankings.

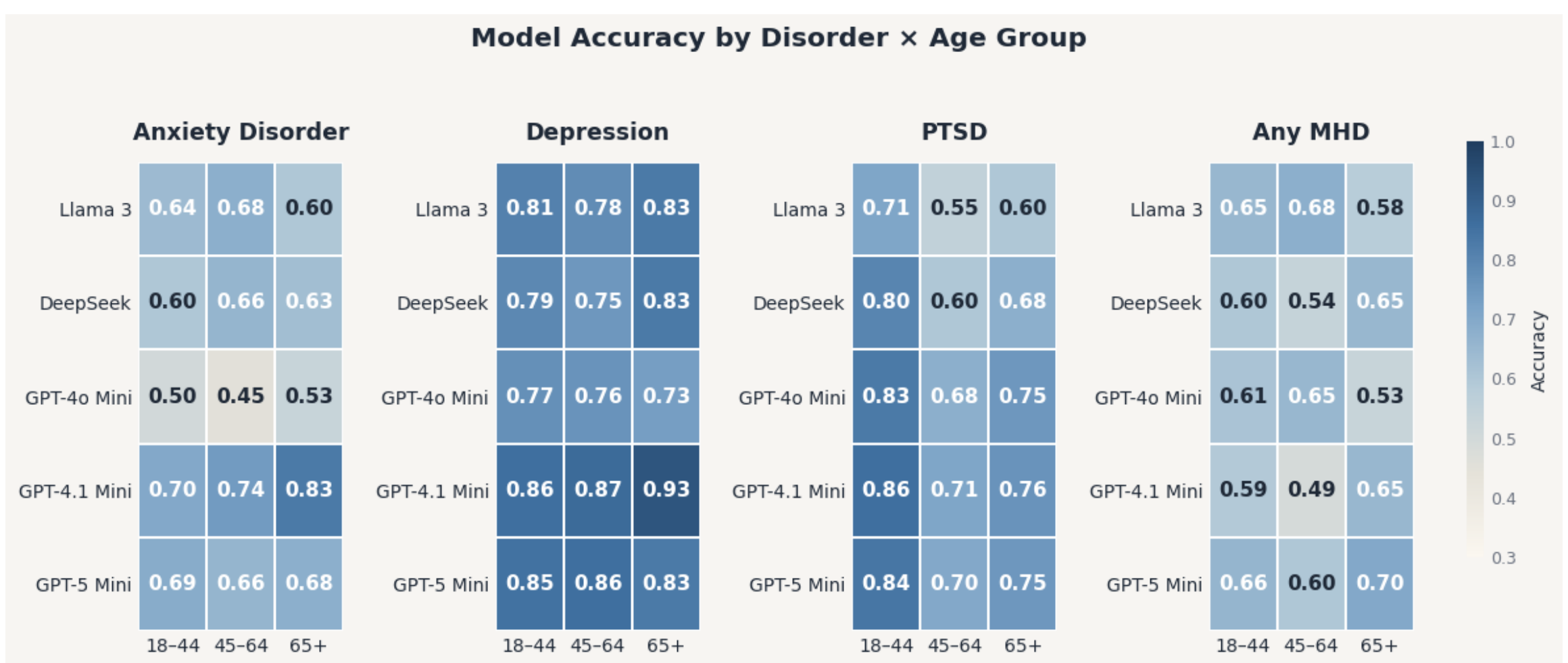

Figure 3. Age-Stratified Model Accuracy Across Mental Health Screening Tasks

Within age strata, Cochran's Q tests indicated significant model-level accuracy differences for most screening tasks. Significant differences were observed for anxiety and depression in all three age groups, for PTSD in the 18–44 and 45–64 groups, and for any current mental health disorder in the 18–44 and 45–64 groups. Differences were not significant for PTSD or any current mental health disorder in the 65+ group. Full age-stratified omnibus and Bonferroni-corrected McNemar results are provided in Appendix B.

**Model Accuracy by Disorder × Race**

|  | Anxiety Disorder White | Anxiety Disorder Non-White | Depression White | Depression Non-White | PTSD White | PTSD Non-White | Any MHD White | Any MHD Non-White |
|---|---|---|---|---|---|---|---|---|
| Llama 3 | 0.65 | 0.56 | 0.81 | 0.74 | 0.65 | 0.65 | 0.64 | 0.65 |
| DeepSeek | 0.63 | 0.48 | 0.78 | 0.73 | 0.73 | 0.69 | 0.59 | 0.52 |
| GPT-4o Mini | 0.50 | 0.42 | 0.76 | 0.68 | 0.77 | 0.77 | 0.61 | 0.65 |
| GPT-4.1 Mini | 0.71 | 0.71 | 0.86 | 0.82 | 0.80 | 0.80 | 0.56 | 0.52 |
| GPT-5 Mini | 0.69 | 0.56 | 0.84 | 0.81 | 0.78 | 0.77 | 0.63 | 0.67 |

Figure 4. Model Accuracy Across Mental Health Screening Tasks by Race Group

Figure 4 summarizes model accuracy across screening tasks stratified by race group. Accuracy ranged from 0.42 to 0.86 across model–task–race combinations. GPT-4.1 Mini showed the highest accuracy for anxiety in both White and Non-White participants (0.71 and 0.71) and for depression among White participants (0.86 and 0.82). While GPT-4.1 Mini and GPT-4o Mini showed the highest PTSD accuracy across race strata. For any current mental health disorder, LLaMA3 and GPT-5 Mini showed comparatively higher accuracy than other models.

Within race strata, Cochran's Q tests indicated significant model-level accuracy differences for all four screening tasks in both White and Non-White participants. Full race-stratified omnibus and Bonferroni-corrected McNemar results are provided in Appendix B. Overall, race-stratified results showed model-dependent variation but no uniform degradation across all tasks in either race group.

## 3.2 Psychiatric Evidence Integration Analysis

### *3.2.1 Evidence Profiles Associated With False-Negative Errors*

To examine evidence-use patterns in GPT-4.1 Mini rationales, inference narratives were coded into three domains: symptom evidence, functional impairment, and protective context. Evidence profiles were compared between true-positive and false-negative cases among SCID-positive participants.

Figure 5 shows that false-negative errors were not uniformly characterized by absence of symptom evidence. For anxiety and PTSD, false-negative cases contained higher symptom-evidence counts than true-positive cases, with the largest difference observed for PTSD. False-negative anxiety cases also contained higher protective-context evidence, suggesting that symptom language was often accompanied by preserved functioning, coping, or social-support cues. In contrast, depression false negatives showed lower functional-impairment evidence than true positives, indicating that depression-positive classifications were more closely aligned with disruption-related reasoning. For any current mental health disorder, true-positive cases contained higher symptom evidence than false negatives, suggesting a more direct symptom-accumulation pattern for broad psychiatric screening.

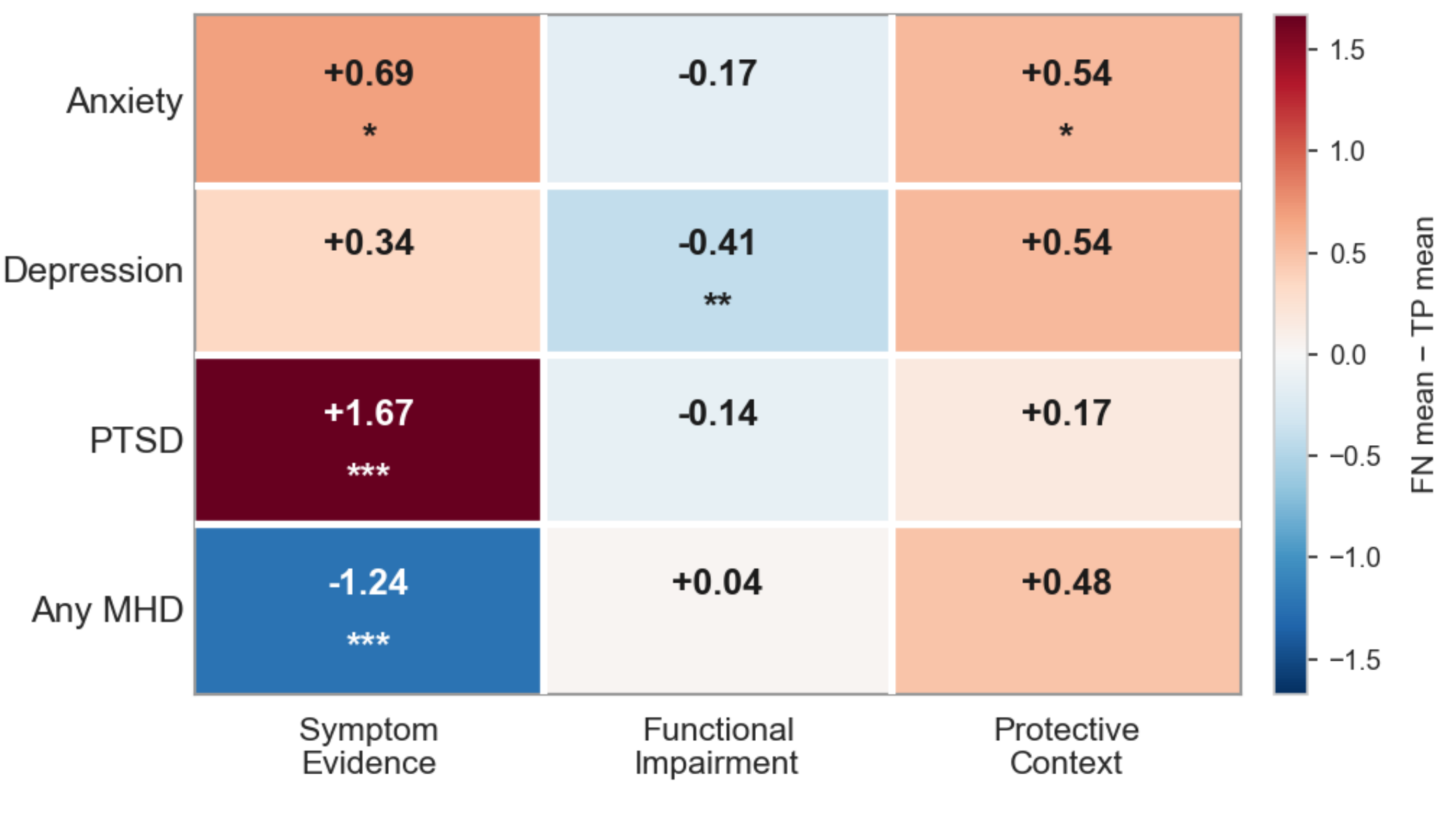

Figure 5. False-Negative Minus True-Positive Evidence Counts Across Screening Tasks

Mann–Whitney U tests with false discovery rate correction supported these task-specific differences. Significant contrasts included higher symptom evidence in false-negative anxiety and PTSD cases, higher protective-context evidence in false-negative anxiety cases, lower functional-impairment evidence in false-negative depression cases, and higher symptom evidence in true-positive any-MHD cases. Full descriptive statistics, effect sizes, and FDR-adjusted p-values are provided in Appendix C.

### 3.2.2 Evidence Features Predicting GPT-4.1 Mini Decisions

Standardized logistic regression models were used to examine how the three evidence domains predicted GPT-4.1 Mini positive outputs and SCID-derived reference-positive labels. As shown in Figure 6A, functional-impairment evidence was positively associated with GPT-4.1 Mini positive outputs across anxiety, depression, and PTSD, whereas protective-context evidence was negatively associated with positive model outputs across tasks. Symptom evidence showed a task-dependent pattern: it was negatively associated with GPT-4.1 Mini positive outputs for anxiety, depression, and PTSD, but positively associated with broad any-MHD classification.

In contrast, the same evidence domains showed weaker associations with SCID-derived reference labels (Figure 6B), with lower AUC values than models predicting GPT-4.1 Mini outputs. This pattern suggests that the coded evidence features more strongly captured GPT-4.1 Mini's decision structure than the underlying SCID-derived diagnostic labels. Full standardized coefficients and model statistics are provided in Appendix C.

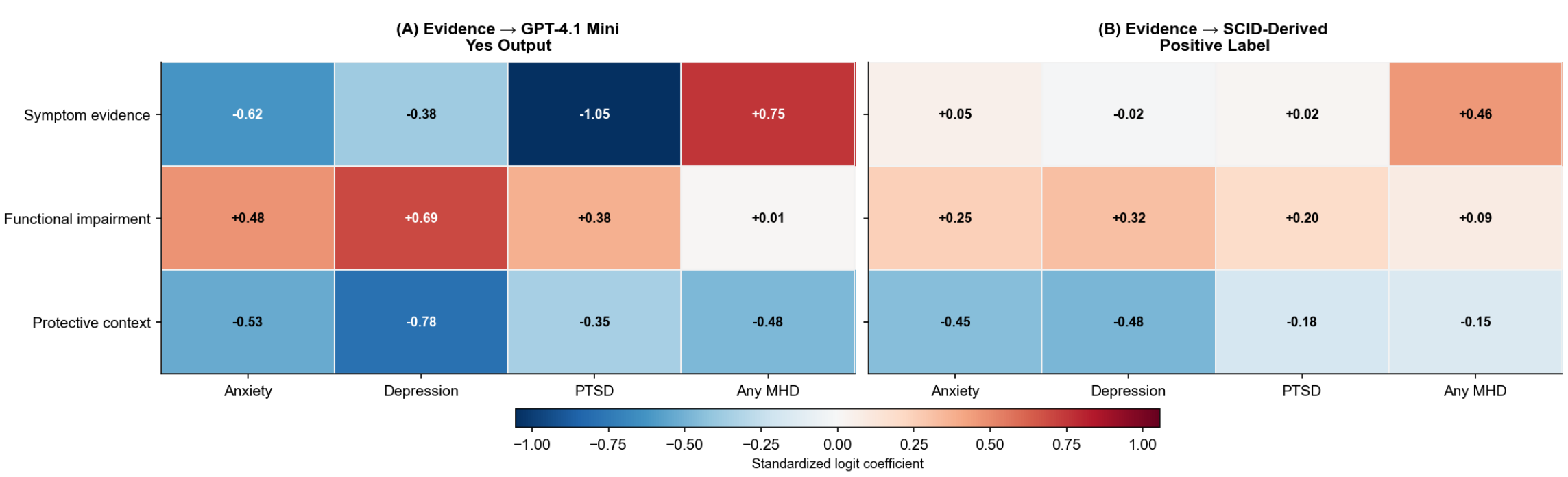

Figure 6. Evidence-Domain Coefficients Predicting Model Outputs and SCID-Derived Labels

### 3.2.3 Participant-Level Evidence Contributions

To visualize participant-level evidence contributions, we generated SHAP-style beeswarm plots from standardized logit contributions derived from the fitted logistic regression models. As shown in Figure 7, participant-level patterns were consistent with the coefficient results in Figure 6. Functional-impairment evidence tended to shift GPT-4.1 Mini outputs toward positive classifications, particularly for depression, whereas protective-context evidence generally shifted outputs away from positive classifications across tasks.

Symptom evidence showed more variable task-specific patterns. For anxiety, depression, and PTSD, high symptom-evidence values did not consistently produce positive logit contributions, whereas symptom evidence contributed more positively to broad any-MHD classification. These participant-level distributions suggest that the evidence-weighting patterns were not driven by isolated cases, but reflected broader decision tendencies across participants. Full contribution plots and supplementary statistics are provided in Appendix C.

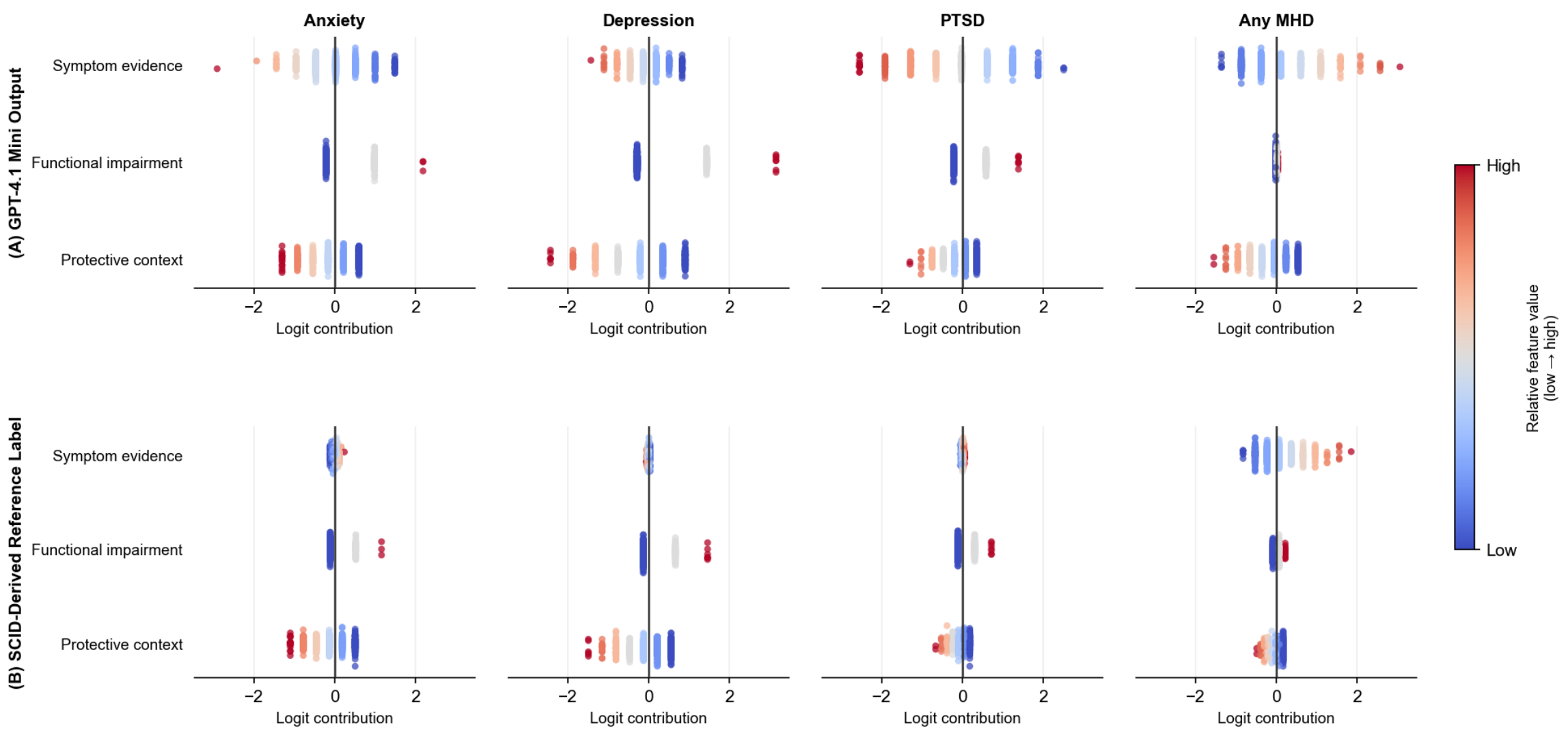

Figure 7 Participant-Level Evidence Contributions to Model Outputs and SCID-Derived Labels

## 4 Discussion

Overall, these findings suggest that LLM-based psychiatric screening is shaped not only by model capability, but also by diagnostic context, subgroup variation, and the way models weight different forms of clinical evidence. Across the three research questions, the results show that stronger overall performance does not necessarily imply uniform reliability across disorders, demographic groups, or error mechanisms. Rather than reflecting a simple failure to detect symptoms, model errors appeared to arise from how symptom evidence was interpreted alongside functional impairment and protective-context cues. This pattern highlights the importance of evaluating LLM screening systems beyond aggregate accuracy, with attention to subgroup fairness and the reasoning patterns that underlie clinically consequential false-negative decisions.

The results indicate that across four psychiatric screening tasks, GPT-4.1 Mini and GPT-5 Mini demonstrated the most consistent overall screening performance, particularly for depression and PTSD classification. This finding is consistent with previous studies that have shown the family models of GPT outperform other models in mental health diagnostic [41,42]. Screening performance varied across diagnostic categories and demographic subgroups, with the clearest variation observed in depression classification, where accuracy was consistently higher for male than female participants across all models. Age- and race-stratified results showed model-dependent variability, but no consistent performance degradation across all diagnostic tasks within any single age or race subgroup. Consistent with our findings, existing studies have similarly reported that LLM-based and language-based mental health screening performance can vary across demographic subgroups, including gender, age, race/ethnicity, and language background, highlighting ongoing concerns regarding fairness and generalizability in AI-assisted psychiatric assessment [43,44].

Although clinical error analysis has long been used to examine misclassification, diagnostic uncertainty, and decision-making failures in mental health assessment [45,46], comparable analyses of reasoning patterns in LLM-based psychiatric screening remain limited [41,47]. Most existing LLM mental health studies focus on overall accuracy or benchmark performance, leaving less clear how models integrate different types of psychiatric evidence when making errors. Our psychiatric evidence integration analysis addresses this gap by showing that GPT-4.1 Mini's false-negative errors were often not caused by an absence of symptom evidence. For anxiety and PTSD, false-negative cases frequently contained substantial symptom-related evidence, suggesting that the model recognized clinically relevant language but did not always convert it into a positive classification. Instead, symptom evidence appeared to be down-weighted when rationales also included preserved functioning, coping behavior, or social-support cues. This pattern was especially evident in PTSD classification, where the model appeared to require stronger convergence between symptoms and functional impairment before assigning a positive label.
Logistic contribution analyses further supported this interpretation: functional-impairment evidence consistently shifted model outputs toward positive classifications, whereas protective-context evidence shifted outputs away from positive classifications. Together, these output-rationale patterns suggest that LLM-based psychiatric screening errors may reflect disorder-specific linguistic evidence-weighting tendencies in model outputs, rather than simple failures of symptom evidence. The findings support research-stage evaluation of LLM screening; but prospective clinical validation, including assessment of real-world sensitivity/specificity tradeoffs and subgroup equity, is required before these systems could be considered for deployment in clinical settings.

Several limitations should be considered. First, all models were evaluated under zero-shot prompting conditions without fine-tuning, clinical calibration, or iterative clinician-style questioning, and results may differ under alternative prompting, adaptation, or multi-turn diagnostic settings. Real-world psychiatric assessment often involves follow-up questions, clarification, and longitudinal context, whereas the present study evaluated single-pass inference from interview text. Second, the output-rational pattern analyses relied on rule-based linguistic operationalizations of symptom evidence, functional impairment, and protective context, which may not capture the full complexity of psychiatric reasoning. Critically, model-generated rationales are output-level text and faithfulness to actual model computations cannot be assumed; these analyses characterize what the model wrote, not how it internally processed the transcript. Third, open-source models may have been affected by context-window constraints, requiring truncation or compression of longer transcripts and potentially losing clinically relevant semantic information.

Future work could examine multi-turn or multi-agent screening frameworks in which models explicitly query uncertain evidence, reconcile symptom and impairment cues, and incorporate reasoning-pattern feedback to improve screening sensitivity and reduce false-negative errors. Recent profile-driven evaluation frameworks further suggest that future LLM psychiatric screening should move beyond aggregate benchmark performance toward construct-level profiles of model behavior [48]. Rather than asking only whether a model produces the correct diagnostic label, future work could characterize models along clinically relevant capability and propensity dimensions, such as symptom recognition, impairment weighting, protective-context sensitivity, uncertainty handling, and follow-up questioning behavior. Such profile-based evaluation may provide a more interpretable and auditable basis for comparing clinical language models and for identifying when a model is likely to miss risk despite recognizing symptom evidence.

## 5 Conclusions

This study evaluated five large language models for zero-shot psychiatric screening using semi-structured stress-focused interviews with SCID-derived diagnostic reference labels. GPT-4.1 Mini and GPT-5 Mini showed the most consistent overall performance, but screening accuracy varied across disorders and demographic subgroups. Evidence-integration analyses further showed that GPT-4.1 Mini's false-negative errors were often not due to absent symptom evidence; rather, symptom cues were frequently down-weighted when accompanied by preserved functioning, coping behavior, or social-support cues, particularly for anxiety and PTSD. The conclusions drawn here may therefore be most applicable to contexts where LLMs are used for single-pass screening from narrative interview text rather than interactive diagnostic assessment. One important limitation suggested by the study is that LLM screening performance may depend not only on symptom recognition, but also on how models weigh competing clinical evidence. Future work should examine calibrated, multi-turn, or multi-agent screening systems that can explicitly resolve uncertainty, integrate symptom and impairment evidence, and improve the transparency and safety of LLM-assisted mental health assessment.

**Ethic Approval and Consent to Participate**

All study procedures were approved by the Kent State University Institutional Review Board (IRB) under the corresponding protocols associated with the included datasets. All participants provided informed consent prior to participation.

## Appendix A. Dataset Description and Interview Materials

*Semi-Structured Experiential Interview Prompts*

| | Baseline | Q1 | Q2 | Q3 | Q4 |
|---|---|---|---|---|---|
| Community Adults With or without psychiatric diagnoses | First, I'd like you to simply tell me about your activities yesterday in detail. For example, tell me what you did from the time you woke up until the time you went to bed? | Please think about the very positive event you identified as occurring when you were about 'X' years old. | Please think about the very negative event you identified as occurring when you were about 'X' years old. | Please think about the event you identified as the time when you felt particularly accepted or supported by others. | Please think about the event you identified as time when you felt particularly rejected or abandoned by others. |

**Example Text from the Semi-Structured Experiential Interview**

| Question1 | Question2 | Question3 | Question4 | Question5 |
|---|---|---|---|---|
| …beginning of the day, uh I have two sons, one is (xxxx) (.) spent the morning with them and uh (.) uh kind of a slow morning (xxxx) um (.) towards the afternoon, uh I started to play outside for a half an hour (.) um (.) spent a little time at the gym, | … being a firefighter has been (.) a challenging and amazing experience, um (.) I've been a firefighter since 2008, I've been a paramedic since 2004 (.) um (.) started in- right after high school so I was 18-years-old, I've enjoyed it, it's been . lot of bad | … I think you're asking why I don't go places, the honest answer is a lot of um. (.) Uh you talk to the people that are around you a little bit who you trust at work but eventually (.) (xxxx) My wife- me and her have been married since we uh-I got on the job….. | . Um I was a newer lieutenant (.) by 3 months first (xxxx) happened (…) uhm (.) first assignment. on it as fast as possible and uhm (..) go inside after (xxxx) after we put most of the fire out um (.)stressful | Uhm (..) two of them that I always think of (xxxx) uh. First one is the first baby that we delivered, uhm (.) early morning call at three, four in the morning heroin (.)…. |

| | experiences uhm. …. | | leading into it….. | |
|---|---|---|---|---|
| This is just daily, or you talkin' about, just when I'm on the fire department? (overlap) Okay, all right Okay (Clears throat) Uh, yesterday, um, came home, had coffee with my wife, talked- did our morning routine with talking, um just about our day, and went about my daily routine kind of just getting stuff done around the house, mowing the lawn, um, just get- taking care of yard work. .. | I think it's turned out being a lot different than what I thought it would be um not so much in a bad way, but I mean I think you get, before you really know the job you get this preconceived notion of what, you know, you're going to be running into a fire or doing some brave thing or you know helping people out, which you do, but not the way I was th- and a lot of the other, | Okay Um coping, coping with certain issues or being a fireman dealing with what we deal with on a day-to-day basis I think that I just more or less find myself sinking into more of my own hobbies and personal interests and ways to kind of escape uh just to get things that I was just doing for the last twenty-four hours out of my head, um and to just focus on something else to kind of get me away. | (whispers) Something else (…) all right How about let's say two years ago(?) um, had a house fire in the middle of the night. (clears throat) Uh we're the second district in so we were the second companies to be there, | Uh, let me think about that one (laughs) .hhh Hhhh (…) It's hard to say cause it's just like your routine things that you do that you know (overlap) (clears throat) Hmm (…) I'm uh- I'm just trying to separate things from just being your routine to something that to me is extraordinary that really sets things apart and I don't think there's really much, |
| Okay. Um. I was 14 and we were on a trip going to Canada. (.) I think- I think it was my first time going there. And um (.) I didn't really want to be (laughs) around my parents for that long 'cause I was young and just, like, eh, but we went. We went to Nia- Niagara Falls,…. | I remember it was (.) a Saturday. Like I was home or whatever and then I think my best friend called me and then she told me that she was moving. | Yeah I'm ready. Yeah I forgot. Okay. Well, I guess I kinda explained to my mom that I have, like, really bad anxiety , like, it's not normal and I (.) actually started going to counseling at White Hall, like, this semester. | Hmm. Okay. Well. Is- yeah a realization. I kinda, like, realized that (.) I have to, like, not interact with him (?). 'Cause we um- we weren't talking for we okay. | So we didn't talk for, like, a month. And then (.) like, he just came out of nowhere and wanted to talk and not be friends (?) but not be with me either. So (.) I, like, kinda, I had to, like, tell him that I didn't wanna do .() what he |

## Appendix Task-Specific Prompt for Depression (MDD)

> You are an AI assisting mental health professionals in identifying signs of major depressive disorder (MDD).
> Instructions:
> 1. Briefly explain (in 2–3 sentences) whether the participant shows signs of depression.
> 2. Then respond with only "Prediction: Yes" or "Prediction: No" on a new line.
> Participant Response:
> "{response_text}"

The output example:

> **LLM inference rationale:**
> The participant does not show prominent signs of depression: they report enjoyment and pride in their work, strong social and family support, effective coping strategies, and positive affect following stressful events. There are no clear reports of persistent low mood, anhedonia, hopelessness, or functional impairment.
> **LLM binary screening output:**
> Prediction: No

## Appendix B. Supplementary Model Performance and Statistical Comparisons

A Cochran's Q test indicated that model accuracy for diagnosing anxiety $\chi^2$ (4) = 95.87, p < .001, depression $\chi^2$ (4) = 48.94, p < .001, ptsd $\chi^2$ (4) = 91.05, p < .001, and any mental health disorders $\chi^2$ (4) = 23.55, p < .001 differed significantly across the five models. Descriptive results showed that GPT4o-Mini had the lowest accuracy for predicting anxiety diagnosis (49%), compared to LLaMA3 (65%), DeepSeek(62%), GPT-4.1 Mini (72%) and GPT-5 Mini(68%). GPT4o-Mini had the lowest accuracy for predicting depression diagnosis (77%), compared to LLaMA3 (80%), DeepSeek(78%), GPT-4.1 Mini (86%) and GPT-5 Mini(85%). LLaMA3 had the lowest accuracy for predicting ptsd diagnosis (65%), compared to DeepSeek(73%), GPT4o-Mini (78%), GPT-4.1 Mini (81%) and GPT-5 Mini(79%). GPT-4.1 Mini had the lowest accuracy for predicting any mental health disorder (MHD) (57%), compared to LLaMA3 (65%), DeepSeek(59%), GPT4o-Mini (62%) and GPT-5 Mini(65%). Post hoc pairwise comparisons McNemar's tests with Bonferroni-correction (α = .005) show specific significant differences between models (See Tables 1 and 2 for full results). Overall, results suggest GPT-4.1 Mini and GPT-5 Miniwere the most consistent in diagnostic accuracy.

**Table B1.** Model Accuracy Proportion Descriptives

| Model | Anxiety | Depression | PTSD | MHD |
|---|---|---|---|---|
| LLaMA3 | .65 | .80 | .65 | .65 |
| Deepseek | .62 | .78 | .73 | .59 |
| GPT4o-Mini | .49 | .77 | .78 | .62 |
| GPT-4.1 Mini | .72 | .86 | .81 | .57 |
| GPT-5 Mini | .68 | .85 | .79 | .65 |

**Table B2.** McNemar Post Hoc Tests of Model Comparisons

| Accuracy Comparisons | Anxiety | Depression | PTSD | *MHD* |
|---|---|---|---|---|
| | $\chi^2$ | $\chi^2$ | $\chi^2$ | $\chi^2$ |
| LLaMA3 vs. Deepseek | 1.43 | 1.30 | 16.49* | 5.92 |
| LLaMA3 vs. GPT4o-Mini | 35.36* | 3.42 | 39.20* | 1.90 |
| LLaMA3 vs. GPT-4.1 Mini | 6.53 | 13.28* | 41.76* | 11.05* |
| LLaMA3 vs. GPT-5 Mini | 1.29 | 7.68 | 39.06* | .05 |
| DeepSeekvs. GPT4o-Mini | 24.67* | .60 | 7.01 | 1.33 |
| DeepSeekvs. GPT-4.1 Mini | 13.91* | 19.96* | 18.13* | 2.17 |
| DeepSeekvs. GPT-5 Mini | 5.59 | 12.76* | 10.56* | 8.76* |
| GPT4o-Mini vs. GPT-4.1 Mini | 53.06* | 29.46* | 2.30 | 5.56 |
| GPT4o-Mini vs. GPT-5 Mini | 50.31* | 22.01* | .22 | 1.65 |
| GPT-4.1 Mini vs. GPT-5 Mini | 4.00 | 2.37 | 2.13 | 22.01* |

Note. * is indicative of a significant difference following Bonferroni correction (α = .005).

A Cochran's Q test indicated that model accuracy of anxiety (females: $\chi^2$ (4) = 14.34, p = .006; males: $\chi^2$ (4) = 106.11, p < .001), depression (females: $\chi^2$ (4) = 31.86, p < .001; males: $\chi^2$ (4) = 37.41, p < .001), ptsd (females: $\chi^2$ (4) = 16.12, p = .003; males: $\chi^2$ (4) = 87.12, p < .001, and any mental health disorder (females: $\chi^2$ (4) = 17.00, p = .002; males: p > .05) diagnoses differed significantly by biological sex across models. Overall, results suggests that GPT-4.1 Mini shows consistently higher diagnostic accuracy across both males and females compared to other models with the exception of any mental health diagnosis group. LLaMA3 shows consistently higher mental health diagnosis accuracy across male and females. Post hoc pairwise comparisons McNemar's tests with Bonferroni-correction (α = .005) show specific significant differences between models (See Tables 3 and 4 for full results).

**Table B3.** Model Accuracy Proportion Descriptives by Sex

| Model | Anxiety | | Depression | | PTSD | | MHD | |
|---|---|---|---|---|---|---|---|---|
| | Male | Female | Male | Female | Male | Female | Male | Female |
| LLaMA3 | .70 | .60 | .89 | .72 | .59 | .71 | .68 | .63 |
| Deepseek | .68 | .56 | .85 | .72 | .72 | .74 | .62 | .56 |
| GPT4o-Mini | .47 | .51 | .79 | .74 | .78 | .78 | .65 | .59 |
| GPT-4.1 Mini | .80 | .64 | .91 | .82 | .84 | .78 | .61 | .53 |
| GPT-5 Mini | .75 | .60 | .87 | .82 | .79 | .80 | .66 | .63 |

**Table B4.** McNemar Post Hoc Tests of Model Comparisons by Sex

| Accuracy Comparisons | Anxiety | Depression | PTSD | MHD |
|---|---|---|---|---|

| | Male $\chi^2$ | Female $\chi^2$ | Male $\chi^2$ | Female $\chi^2$ | Male $\chi^2$ | Female $\chi^2$ | Male $\chi^2$ | Female $\chi^2$ |
|---|---|---|---|---|---|---|---|---|
| LLaMA3 vs. Deepseek | .45 | .79 | 3.23 | .00 | 16.75* | 1.29 | 2.42 | 3.15 |
| LLaMA3 vs. GPT4o-Mini | 33.08* | 5.82 | 14.58* | .54 | 34.22* | 6.61 | .68 | 1.01 |
| LLaMA3 vs GPT-4.1 Mini | 8.89* | .72 | 1.33 | 12.29* | 44.00* | 4.63 | 4.20 | 6.40 |
| LLaMA3 vs. GPT-5 Mini | 2.22 | .01 | .35 | 16.00* | 32.58* | 7.93* | .21 | .00 |
| DeepSeekvs. GPT4o-Mini | 28.01* | 2.25 | 3.94 | .51 | 4.34 | 2.22 | .44 | .74 |
| DeepSeekvs. GPT-4.1 Mini | 12.10* | 3.22 | 7.61 | 11.46* | 18.48* | 2.42 | .49 | 1.42 |
| DeepSeekvs. GPT-5 Mini | 5.16 | 1.10 | .84 | 14.79* | 5.56 | 4.36 | 1.69 | 7.52 |
| GPT4o-Mini vs. GPT-4.1 Mini | 53.77* | 8.17* | 25.69* | 6.69 | 3.84 | .00 | 1.59 | 3.94 |
| GPT40Mini vs. GPT-5 Mini | 50.25* | 6.97 | 12.60* | 8.82* | .00 | .30 | .06 | 2.33 |
| GPT-4.1 Mini vs. GPT-5 Mini | 3.25 | .928 | | | 6.5 | | 5.03 | 17.42* |

Note. * is indicative of a significant difference following Bonferroni correction (α = .005).

A Cochran's Q test indicated that model accuracy of anxiety (white: $\chi^2$ (4) = 64.58, $p < .001$; non-white: $\chi^2$ (4) = 20.63, $p < .001$), depression (white: $\chi^2$ (4) = 37.76, $p < .001$; non-white: $\chi^2$ (4) = 10.88, $p = .028$), ptsd (white: $\chi^2$ (4) = 66.06, $p < .001$; non-white: $\chi^2$ (4) = 12.41, $p = .015$), and any mental health disorder (white: $\chi^2$ (4) = 14.90, $p = .005$; non-white: $\chi^2$ (4) = 11.38, $p = .023$) diagnoses differed significantly by race across models. Overall, results suggests that GPT-4.1 Mini shows consistently higher diagnostic accuracy across both white and non-white individuals compared to other models with the exception to any mental health diagnosis group. LLaMA3 and GPT-5 Minishow consistently higher mental health diagnosis accuracy across white and non-white individuals. Post hoc pairwise comparisons McNemar's tests with Bonferroni-correction (α = .005) show specific significant differences between models (See Tables 5 and 6 for full results).

**Table B5.** Model Accuracy Proportion Descriptives by Race

| Model | Anxiety | | Depression | | PTSD | | MHD | |
|---|---|---|---|---|---|---|---|---|
| | White | Non-White | White | Non-White | White | Non-White | White | Non-White |
| LLaMA3 | .65 | .56 | .81 | .74 | .65 | .65 | .64 | .65 |
| Deepseek | .63 | .48 | .78 | .73 | .73 | .69 | .59 | .52 |
| GPT4o-Mini | .50 | .42 | .76 | .68 | .77 | .77 | .61 | .65 |
| GPT-4.1 Mini | .71 | .71 | .86 | .82 | .80 | .80 | .56 | .52 |
| GPT-5 Mini | .69 | .56 | .84 | .81 | .78 | .77 | .63 | .67 |

**Table B6.** McNemar Post Hoc Tests of Model Comparisons by Race

| Accuracy Comparisons | Anxiety | | Depression | | PTSD | | MHD | |
|---|---|---|---|---|---|---|---|---|
| | White $\chi^2$ | Non-White $\chi^2$ | White $\chi^2$ | Non-White $\chi^2$ | White $\chi^2$ | Non-White $\chi^2$ | White $\chi^2$ | Non-White $\chi^2$ |
| LLaMA3 vs. Deepseek | .60 | | 1.70 | | 13.60* | | 2.88 | 2.56 |
| LLaMA3 vs. GPT4o-Mini | 25.44* | 3.03 | 3.21 | | 28.46* | | 1.12 | .00 |
| LLaMA3 verses GPT-4.1 Mini | 3.29 | 3.89 | 9.14* | | 30.62* | | 8.06* | 2.33 |
| LLaMA3 vs. GPT-5 Mini | 1.23 | .00 | 4.83 | | 28.77* | 4.32 | .19 | .00 |
| DeepSeekvs. GPT4o-Mini | 18.45* | .46 | .33 | .35 | 3.61 | | .45 | 3.45 |
| DeepSeekvs. GPT-4.1 Mini | 6.98 | 9.03* | 15.71* | | 11.69* | | 3.21 | |
| DeepSeekvs. GPT-5 Mini | 3.70 | | 9.59* | | 5.96 | | 2.61 | |
| GPT4o-Mini vs. GPT-4.1 Mini | 34.44* | 10.87 * | 21.78* | | 1.92 | | 4.52 | |

| | | | | | |
|---|---|---|---|---|---|
| GPT40Mini vs. GPT-5 Mini | 37.79* | 3.56 | 16.28 | .16 | .39 |
| GPT-4.1 Mini vs. GPT-5 Mini | 1.12 | 4.65 | | 1.64 | 13.57* |

Note. * is indicative of a significant difference following Bonferroni correction (α = .005).

A Cochran's Q test indicated that model accuracy of anxiety (18-44: $\chi^2$ (4) = 46.66, p < .001; 45-64: $\chi^2$ (4) = 48.81, p < .001; 65+: $\chi^2$ (4) = 11.11, p = .025), depression (18-44: $\chi^2$ (4) = 23.61, p < .001; 45-64: $\chi^2$ (4) = 19.81, p < .001; 65+: $\chi^2$ (4) = 12.80, p = .012), ptsd (18-44: $\chi^2$ (4) = 63.50, p < .001; 45-64: $\chi^2$ (4) = 23.91, p < .001; 65+: p > .05), and any mental health disorder (18-44: $\chi^2$ (4) = 10.05, p = .040; 45-64: $\chi^2$ (4) = 26.31, p < .001; 65+: p > .05) diagnoses differed significantly by age across models. Overall, results suggests that GPT-4.1 Mini shows consistently higher diagnostic accuracy across all age groups compared to other models with the exception to any mental health diagnosis group. GPT-5 Minishows consistently higher mental health diagnosis accuracy across all age groups. Post hoc pairwise comparisons McNemar's tests with Bonferroni-correction (α = .005) show specific significant differences between models (See Tables 7 and 8 for full results).

**Table B7.** Model Accuracy Proportion Descriptives by Age

| Model | Anxiety | | | Depression | | | PTSD | | | MHD | | |
|---|---|---|---|---|---|---|---|---|---|---|---|---|
| | 18-44 | 45-64 | 65+ | 18-44 | 45-64 | 65+ | 18-44 | 45-64 | 65+ | 18-44 | 45-64 | 65+ |
| LLaMA3 | .64 | .68 | .60 | .81 | .78 | .83 | .71 | .55 | .60 | .65 | .68 | .58 |
| Deepseek | .60 | .66 | .63 | .79 | .75 | .83 | .80 | .60 | .68 | .60 | .54 | .65 |
| GPT4o-Mini | .50 | .45 | .53 | .77 | .76 | .73 | . 83 | .68 | .75 | .61 | .65 | .53 |
| GPT-4.1 Mini | .70 | .74 | .83 | .86 | .87 | .93 | .86 | .71 | .76 | .59 | .49 | .65 |
| GPT-5 Mini | .69 | .66 | .68 | .85 | .86 | .83 | .84 | .70 | .75 | .66 | .60 | .70 |

**Table B8.** McNemar Post Hoc Tests of Model Comparisons by Age

| Accuracy Comparisons | Anxiety | | | Depression | | | PTSD | | | MHD | | |
|---|---|---|---|---|---|---|---|---|---|---|---|---|
| | 18-44 $\chi^2$ | 45-64 $\chi^2$ | 65+ $\chi^2$ | 18-44 $\chi^2$ | 45-64 $\chi^2$ | 65+ $\chi^2$ | 18-44 $\chi^2$ | 45-64 $\chi^2$ | 65+ $\chi^2$ | 18-44 $\chi^2$ | 45-64 $\chi^2$ | 65+ $\chi^2$ |
| LLaMA3 vs. Deepseek | 1.62 | .03 | | .80 | .30 | | 16.36* | 1.42 | | 1.80 | 7.35 | |
| LLaMA3 vs. GPT4o-Mini | 16.99* | 20.13* | | 2.41 | .09 | | 25.57* | 9.59* | | 1.26 | .29 | |
| LLaMA3 vs. GPT-4.1 Mini | 2.41 | 1.40 | | 4.00 | 7.26 | | 30.53* | 8.68* | | 3.11 | 13.43* | |
| LLaMA3 vs. GPT-5 Mini | 1.62 | .02 | | 2.42 | | | 25.14* | 10.47* | | .04 | 2.24 | |
| DeepSeekvs. GPT4o-Mini | 8.20* | 21.12* | | .47 | .00 | | 2.42 | 3.06 | | .05 | 4.49 | |
| DeepSeekvs. GPT-4.1 Mini | 7.82* | 2.09 | | 8.82* | 7.90* | | 8.80* | 6.92 | | .36 | | |
| DeepSeekvs. GPT-5 Mini | 6.88 | .00 | | 6.35 | 6.24 | | 3.25 | 6.62 | | 5.35 | 2.53 | |
| GPT4o-Mini vs. GPT-4.1 Mini | 23.38* | 25.54* | | 14.79* | 8.03* | | 2.22 | .19 | | .66 | 11.16* | |
| GPT40Mini vs. GPT-5 Mini | 28.06* | 20.42* | | 11.16* | 7.50 | | .03 | .15 | | 3.02 | .77 | |
| GPT-4.1 Mini vs. GPT-5 Mini | .22 | 3.03 | | | | | | | | 10.30* | 11.17* | |

Note. * is indicative of a significant difference following Bonferroni correction (α = .005)

## Appendix C. Inference-Text Keywords and Supplementary Post-hoc Results

This appendix reports the validated keyword lists and supplementary post-hoc analyses used to characterize GPT-4.1 Mini inference text under task-specific and unified prompting. Symptom keywords were derived from PHQ-9, GAD-7, and PCL-5 item or cluster content. Severity-related terms were exploratory and are reported as null supplementary analyses.

### Table C1. Rule-Based Evidence Coding Dictionaries

| Evidence Category | Subdomain | Representative Keywords / Expressions |
|---|---|---|
| Symptom Evidence | Anxiety-related symptoms | anxiety, anxious, worry, panic attack, fear, avoidance, restless, on edge, irritability, compulsive, obsessive, social anxiety |
| Symptom Evidence | Depression-related symptoms | depressed, sadness, low mood, hopelessness, anhedonia, loss of interest, fatigue, tired, low energy, insomnia, appetite change, worthlessness, guilt, suicidal ideation |
| Symptom Evidence | PTSD-related symptoms | PTSD, trauma, traumatic, flashbacks, intrusive memories, nightmares, hypervigilance, easily startled, emotional numbing, detachment |
| Symptom Evidence | General mental health indicators | mental health, psychiatric, diagnosis, self-harm, addiction, psychosis, hallucination, delusion, mania, bipolar |
| Functional Impairment | Daily functioning disruption | impaired, interference, difficulty functioning, unable to, cannot, disruption, affected work, affected school, occupational impairment, social impairment |
| Functional Impairment | Coping/functioning difficulties | struggling to function, struggling to work, difficulty concentrating, difficulty sleeping, difficulty coping |
| Preserved Functioning | Maintained functioning | functioning well, no significant impairment, maintains work, maintains responsibilities, able to function, normal functioning, daily routine |
| Coping Evidence | Active coping strategies | cope, coping strategies, stress management, relaxation, therapy, therapist, counseling, mindfulness, meditation |
| Social Support | Interpersonal support | support system, supportive, family, friends, partner, spouse, connected, spending time with others, not isolated |

The rule-based evidence dictionaries were designed as transparent operationalizations of recurring reasoning patterns observed in GPT-4.1 Mini inference narratives. Dictionaries were informed by prior computational mental health research using keyword-based linguistic markers and psychiatric symptom lexicons. These dictionaries were used exclusively for computational narrative analysis and were not intended as diagnostic instruments.

### Table C2. Mann–Whitney TP vs FN Comparisons

| **Task** | **Evidence Metric** | **TP Mean** | **FN Mean** | **FN − TP** | **Rank-Biserial r** | **p_FDR** |
|---|---|---|---|---|---|---|
| Anxiety | Symptom evidence | 3.52 | 4.20 | +0.69 | 0.33 | .012* |
| Anxiety | Functional impairment | 0.39 | 0.22 | −0.17 | −0.18 | .078 |
| Anxiety | Protective context | 0.74 | 1.28 | +0.54 | 0.27 | .039* |
| Depression | Symptom evidence | 3.35 | 3.69 | +0.34 | 0.07 | .626 |
| Depression | Functional impairment | 0.57 | 0.16 | −0.41 | −0.34 | .005** |
| Depression | Protective context | 0.78 | 1.33 | +0.54 | 0.19 | .250 |
| PTSD | Symptom evidence | 4.67 | 6.34 | +1.67 | 0.53 | <.001*** |
| PTSD | Functional impairment | 0.46 | 0.32 | −0.14 | −0.10 | .447 |
| PTSD | Protective context | 1.00 | 1.17 | +0.17 | 0.13 | .383 |
| Any MHD | Symptom evidence | 4.07 | 2.83 | −1.24 | −0.39 | <.001*** |
| Any MHD | Functional impairment | 0.56 | 0.60 | +0.04 | 0.05 | .530 |
| Any MHD | Protective context | 1.31 | 1.79 | +0.48 | 0.16 | .090 |

**Table C3. Logistic Regression Coefficients Predicting GPT-4.1 Mini Positive Outputs**

| Task | Evidence Feature | β | AUC | N |
|---|---|---|---|---|
| Anxiety | Symptom evidence | −0.62 | 0.75 | 555 |
| Anxiety | Functional impairment | +0.48 | 0.75 | 555 |
| Anxiety | Protective context | −0.53 | 0.75 | 555 |
| Depression | Symptom evidence | −0.38 | 0.76 | 555 |
| Depression | Functional impairment | +0.69 | 0.76 | 555 |
| Depression | Protective context | −0.78 | 0.76 | 555 |
| PTSD | Symptom evidence | −1.05 | 0.78 | 554 |
| PTSD | Functional impairment | +0.38 | 0.78 | 554 |
| PTSD | Protective context | −0.35 | 0.78 | 554 |
| Any MHD | Symptom evidence | +0.75 | 0.73 | 553 |
| Any MHD | Functional impairment | +0.01 | 0.73 | 553 |
| Any MHD | Protective context | −0.48 | 0.73 | 553 |

**Table C4. Logistic Regression Coefficients Predicting SCID-Derived Positive Labels**

| Task | Evidence Feature | β | AUC | N |
|---|---|---|---|---|
| Anxiety | Symptom evidence | +0.05 | 0.62 | 555 |
| Anxiety | Functional impairment | +0.25 | 0.62 | 555 |
| Anxiety | Protective context | −0.45 | 0.62 | 555 |
| Depression | Symptom evidence | −0.02 | 0.60 | 555 |
| Depression | Functional impairment | +0.32 | 0.60 | 555 |
| Depression | Protective context | −0.48 | 0.60 | 555 |
| PTSD | Symptom evidence | +0.02 | 0.55 | 554 |
| PTSD | Functional impairment | +0.20 | 0.55 | 554 |
| PTSD | Protective context | −0.18 | 0.55 | 554 |
| Any MHD | Symptom evidence | +0.46 | 0.62 | 553 |
| Any MHD | Functional impairment | +0.09 | 0.62 | 553 |
| Any MHD | Protective context | −0.15 | 0.62 | 553 |